\documentclass[letterpaper, 10 pt, conference]{ieeeconf}  % Comment this line out if you need a4paper

\IEEEoverridecommandlockouts                              % This command is only needed if
\usepackage{graphics} % for pdf, bitmapped graphics files
\usepackage{epsfig} % for postscript graphics files
\usepackage{amsmath} % assumes amsmath package installed
\usepackage{amssymb}  % assumes amsmath package installed
\usepackage{multirow}
\usepackage{xcolor}
\usepackage[ruled,vlined,linesnumbered]{algorithm2e}

\newcommand{\argmin}{\mathop{\mathrm{arg\,min}}\nolimits}

\newlength\mylen

\title{\LARGE \bf
MSTC${}^\ast$:Multi-robot  Coverage  Path  Planning under Physical Constraints}

\author{Jingtao Tang, Chun Sun and Xinyu Zhang% stops a space
\thanks{The authors are with Shanghai Key Laboratory of Trustworthy Computing, Engineering Research Center of Software/Hardware Co-design Technology and Application (MoE) and School of Software Engineering, East China Normal University, Shanghai. Xinyu~Zhang is the corresponding author. E-mail: {\small \{xyzhang\}@sei.ecnu.edu.cn}.
}%
}

\begin{document}

\maketitle
\thispagestyle{empty}
\pagestyle{empty}

\begin{abstract}
For large-scale tasks, coverage path planning (CPP) can benefit greatly from multiple robots.
In this paper, we present an efficient algorithm MSTC${}^\ast$ for multi-robot coverage path planning (mCPP) based on spiral spanning tree coverage (Spiral-STC).
Our algorithm incorporates strict physical constraints like terrain traversability and material load capacity.
We compare our algorithm against the state-of-the-art in mCPP for regular grid maps and real field terrains in simulation environments. 
The experimental results show that our method significantly outperforms existing spiral-STC based mCPP methods.
Our algorithm can find a set of well-balanced workload distributions for all robots and therefore, achieve the overall minimum time to complete the coverage.
\end{abstract} 
\section{Introduction} \label{sec:intro}

% introduction and background of the problem %

Coverage path planning (CPP) is the problem of determining a set of paths that cover the area of interest while avoiding obstacles\cite{CPPsurvey2013}. Coverage path planning has many indoor and outdoor robotic applications, such as vacuum cleaning robots\cite{vacuum-cleaner}, autonomous underwater vehicles\cite{underwater-vehicles}, unmanned aerial vehicles\cite{aerial-vehicles}, demining robots\cite{demining-robots}, automated harvesters\cite{automated-harvesters}, planetary exploration\cite{planetary-exploration}, search and rescue operations\cite{search-rescue}, lawn mowers\cite{cpp-lawn-mower}, massive afforestation\cite{afforestation}.

Coverage path planning has been received a lot of attention in robotics and there are a considerable research work addressing this problem\cite{CPPsurvey2013}. This includes cellular decomposition\cite{cpp-boustrophedon, cpp-trapezoid, cpp-morse-decomp}, gird map\cite{grid-boustrophedon}, spanning tree coverage\cite{SpiralSTC}, neural network-based coverage\cite{cpp-NN-1, cpp-nn-2}, graph-based coverage\cite{cpp-graph}, optimal coverage\cite{cpp-optimal-1, cpp-optimal-2}, coverage under uncertainty\cite{cpp-uncertainty-1, cpp-uncertainty-2}. Most these approaches were designed mainly for a single robot.

For large-scale tasks, coverage path planning can benefit greatly from multi-robot systems. First, a multi-robot system clearly completes the task fast due to workload distribution. Second, multiple robots can collaborate with each other to accomplish complex tasks efficiently. Third, multi-robots improve robustness in case of failure of some robots.
Though there are many advantages using multiple robots, the
research in multi-robot coverage path planning is relatively limited since some extra factors (e.g., data sharing, complex path generation, task division/allocation, physical constraints, etc.) need to be taken into account. Many approaches extended single-robot algorithms to handle multi-robot systems using workload division/distribution\cite{survey-mcpp}.

Despite the exciting potential applications, designing a scalable and practical multi-robot coverage path planning (mCPP) algorithm under strict physical constraints remains a challenging problem. Our research was developed from an ambitious program for tree planting robots to restore vast degraded lands. These terrains may exhibit complex surface and topology. The robots have limited energy and workload/material capacity (e.g., 100 tree saplings per load for planting robots and 500kg water per load for watering robots). Coverage path planning with physical constraints is a relatively new topic and energy constraints were often considered in limited research literatures\cite{CPP-GA, online-cpp-battery-powered, CPP-energy-constraint, fermat-spiral}.

\textbf{Main Results:} We propose a novel method namely MSTC${}^\ast$ (Multi-robot Spanning Tree Coverage Star), to solve the mCPP problem under physical constraints of traversability and limited workload/material capacity. We treat mCPP as the problem of partitioning a topological loop and assign each partition to one robot. To find a set of well-balanced partitions, we start with a set of naïve 
partitions and iteratively generate balanced partitions for all robots by minimizing the maximum weights. Our balanced-MSTC${}^\ast$ uses the strategy of balanced cut to search the most unbalanced two partitions (with the maximum and minimum weights, respectively). This strategy is a greedy algorithm and is able to gradually approximate the optimal partitions. Our algorithm can find a set of well-balanced workload distributions for all robots and therefore, achieve the overall  optimal  time  to  complete  the  coverage.
We compare our algorithm against other spiral-STC based mCPP methods on regular grid maps and real field terrains in simulation environments. The results show that our MSTC${}^\ast$ algorithm outperforms the state-of-the-art, like classic MSTC (MSTC-NB)\cite{MSTC}, MSTC with backtracking (MSTC-BO)\cite{MSTC} and Multi-robot Forest Coverage (MFC)\cite{MFC}. 
\section{Related Work} \label{sec:related}
Coverage path planning is well studied for a single robot and we refer readers to~\cite{grid-boustrophedon,CPPsurvey2013,survey-mcpp} for extensive survey. Here, we briefly review the literature relevant to our work.

Our method is inspired by spiral spanning tree coverage (Spiral-STC)~\cite{SpiralSTC} for a single robot and multi-robot spanning tree coverage (MSTC)~\cite{MSTC} for unweighted graph. The latter improved the efficiency and robustness by introducing redundancy and backtracking optimization using multiple robots. Agmon et al. constructed a spanning tree by minimizing the time to complete the coverage~\cite{st-opt-mstc-1, st-opt-mstc-2}.
Zheng et.al. proposed multi-robot forest coverage (MFC) to solve the mCPP problem using
multiple minimal spanning trees to cover the terrain generated by min-max tree cover algorithm~\cite{min-max-tree-cover}. The algorithm works for both unweighted terrains~\cite{MFC} and weighted terrains~\cite{MFC-weighted, MFC-2010}.
Most algorithms assumed that robots can fully cover the environment without recharging or refilling. However, in the real-world applications, many physical constraints need to be considered. Coverage path planning with physical constraints is a relatively new topic. In some recent work\cite{CPP-GA, online-cpp-battery-powered, CPP-energy-constraint, fermat-spiral}, energy limitations were considered on coverage path planning for a single robot.
%Yet there are few mCPP researches for planar robots under the constraints of energy, workload capacity and other physical constraints,
Moreover, Sipahioglu et al. proposed a generalized Voronoi
diagram based method to solve the problem of mCPP under energy capacity\cite{mcpp-gvd-energy}. Huang et al. developed a quadtree and spiral-STC based method to adapt mCPP to different land types\cite{mcpp-land-cover}. Our algorithm incorporates strict physical constraints like terrain traversability and material load capacity.
%Our work considers to balance the workload (i.e. energy) of each robots and further improve the efficiency of mCPP, under a limited workload capacity and terrain traversability constraint.

%Although recently there are works intended to solve the coverage path planning  problem concerning the energy efficiency or constraints~\cite{CPP-GA, online-cpp-battery-powered, CPP-energy-constraint, fermat-spiral}, there exists few works on solving the MCPP problem under limited material capacity constraint.

\newcommand{\node}{\mbox{$\pi$}\xspace}
\newcommand{\key}{\mbox{$\rho$}\xspace}

\section{Problem Definition \& Preliminaries}\label{sec:problem}
\subsection{Problem Definition}\label{sec:definition}
\begin{figure}[!thb]
\centering
\includegraphics[width=\linewidth]{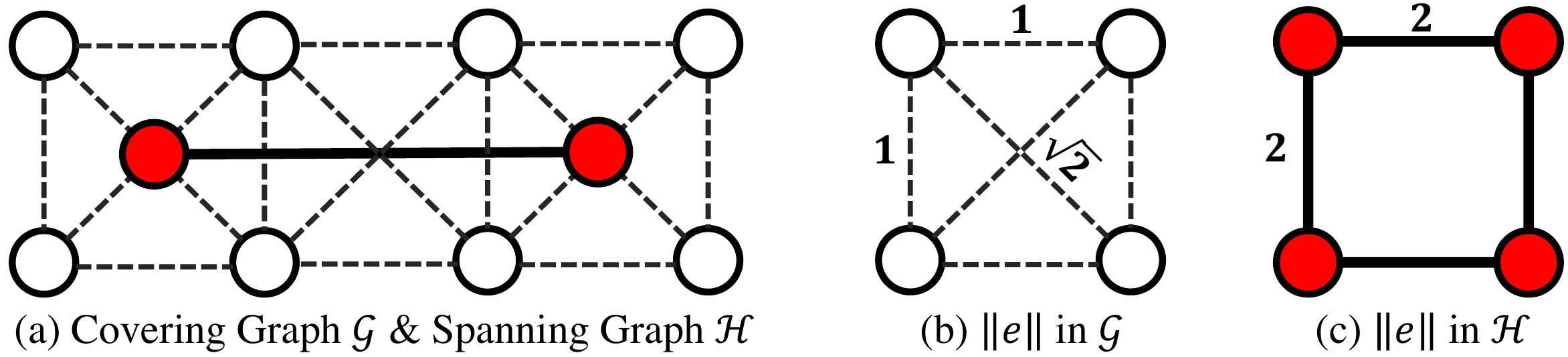}
\caption{Covering graph $\mathcal{G}$ is represented by white nodes and dotted edges, and spanning graph $\mathcal{H}$ is represented by red nodes and solid edges. (a) A spanning node is generated from four adjacent covering nodes; (b) A weight (1 or $\sqrt{2}$) is assigned to the edges of $\mathcal{G}$; (c) A weight 2 is assigned to the edges of $\mathcal{H}$.}
\label{fig:demo_edges}
\end{figure}

The goal of mCPP is to cover a given terrain using multiple robots. The terrain is divided into a large number of cells and the cell dimensions depend on specific applications. For example, in our program for tree planting, the spacing between the lines in plantation is 5 meters and the spacing of plants within a line is 3 meters. Then we represent the terrain as a graph, namely \emph{covering graph}, denoted by $\mathcal{G}$ (see Fig.~\ref{fig:demo_edges}-(a)). The nodes in $\mathcal{G}$ is covering nodes, denoted by $\node$. Two adjacent nodes are connected by an edge $e$. In spiral-STC based algorithms, a spanning graph is used to efficiently generate coverage paths. Here, we denote a spanning graph by $\mathcal{H}$ and its nodes are \textit{spanning nodes}. Note that a covering node is associated to only one spanning node.
Both $\mathcal{G}$ and $\mathcal{H}$ are edge-weighted graphs. The edge weights $\|e\|$ are shown in Fig.~\ref{fig:demo_edges}.
\vspace{-1em}
\vspace{-1em}

\begin{figure}[!thb]
\centering
\includegraphics[width=0.90\linewidth]{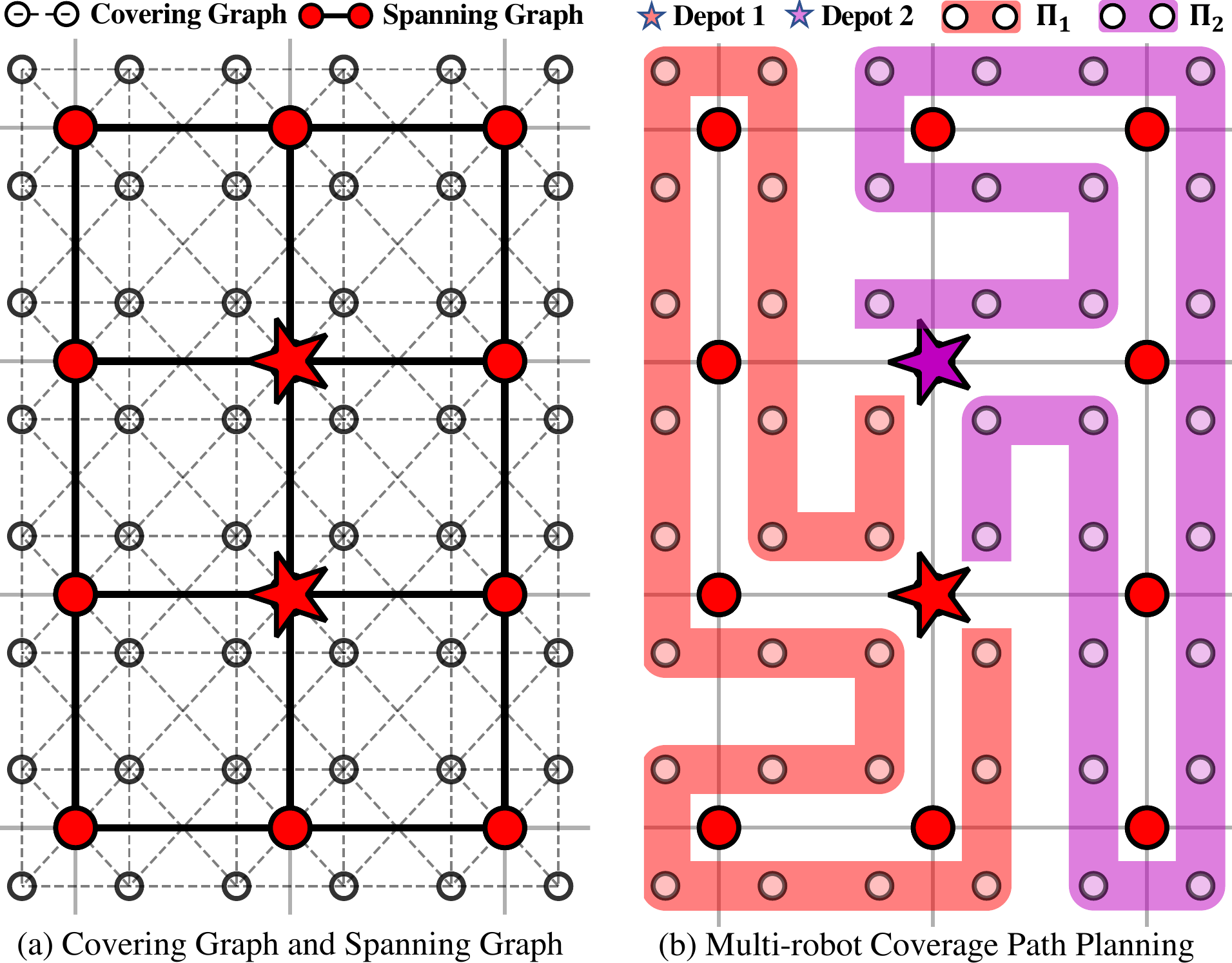}
\caption{Multi-robot coverage path planning using our method MSTC${}^\ast$. (a) Covering graph $\mathcal{G}$ and spanning graph $\mathcal{H}$ constructed from a given terrain; (b) Coverage planning results using two robots starting from their individual depots (highlighting with stars). Their coverage paths are highlighted in red and purple, respectively. Note that, a spanning graph is used to efficiently generate coverage paths.}
\label{fig:demo_2_nodes}
\end{figure}

Given $\mathcal{G}$ and $k$ robots, a coverage path $\Pi_i$ travelled by robot $\mathcal{R}_i$ is a set of nodes $\{\node_i^j\}$ in $\mathcal{G}$ (see Fig.~\ref{fig:demo_2_nodes}). 
For robot $\mathcal{R}_i$, its accumulating weight (cost) is  denoted by $\mathcal{W}_{\Pi_i}$ .
We aim at computing a set of coverage paths $\{\Pi_1,\Pi_2,\ldots,\Pi_k\}$ for $k$ robots through minimizing the maximum of $\mathcal{W}_{\Pi_i}$.
Then the problem can be formulated as follows.
\begin{align}\label{eqn:mcpp_task}
\argmin_{\{\Pi_i\}}\limits\,\left( \max_{1\leq i\leq k}\limits \left(\mathcal{W}_{\Pi_i}\right)\right)
%s.t. \; \mathcal{N} = \bigcup_{\,i=0}^{\,k-1}\nolimits \,\{x_j^i\,|\,j\in [0, |\pi_i|)\}
\end{align}
Besides requirements described in \cite{CPPsurvey2013}, the mCPP problem in this paper has the following constraints:
\begin{itemize}
\item \emph{Depots}: a robot $\mathcal{R}_i$ has its own depot, denoted by $\node^d_i$.
\item \emph{Cover and return}: a robot returns to its individual depot when its tasks are accomplished.
\item \emph{Workload capacity}: a robot has limited material capacity per load and need to immediately return to its depot to refill when its material runs out.
\end{itemize}
\vspace{-1em}

\subsection{Spiral-STC for A Single Robot}
In this section, we introduce the spiral-STC algorithm that inspires our work. Spiral-STC was originally proposed for a single robot. Given a covering graph $\mathcal{G}$ and its associating spanning tree $\mathcal{H}$, spiral-STC performs a counter-clockwise depth-first-search in $\mathcal{H}$ and then generates a circumnavigating coverage path in $\mathcal{G}$ by following the right-side of traversal route (i.e., ordered spanning tree edges) in $\mathcal{H}$.

%coverage path planning problem on covering graph $\mathcal{G}$.
%For convenience, we denote the coverage path generated by Spiral-STC on $\mathcal{G}$ as $\Pi$.

%Assuming that $\mathcal{G}$ and $\mathcal{H}$ are unweighted graphs, the original Spiral-STC algorithm solves the single robot coverage path planning problem by performing a Counter-ClockWise (CCW) Depth-First-Search (DFS) on $\mathcal{H}$, and then generating a circumnavigating coverage path on $\mathcal{G}$, by always following the right-side of the traversal route (i.e. ordered spanning tree edges) on $\mathcal{H}$.
%The pseudo-code of Spiral-STC in Alg.~\ref{alg:Spiral-STC} describes how the recursive DFS procedure is executed, note that the traversal order is set to CCW in line 9.
%As long as the traversal does not revisit same spanning nodes of $\mathcal{H}$, the correctness of complete coverage on $\mathcal{G}$ using Spiral-STC algorithm can be easily proven using the Euler's Circuit theorem, since each node has even degrees.

\begin{figure}[!thb]
\centering
\includegraphics[width=0.9\linewidth]{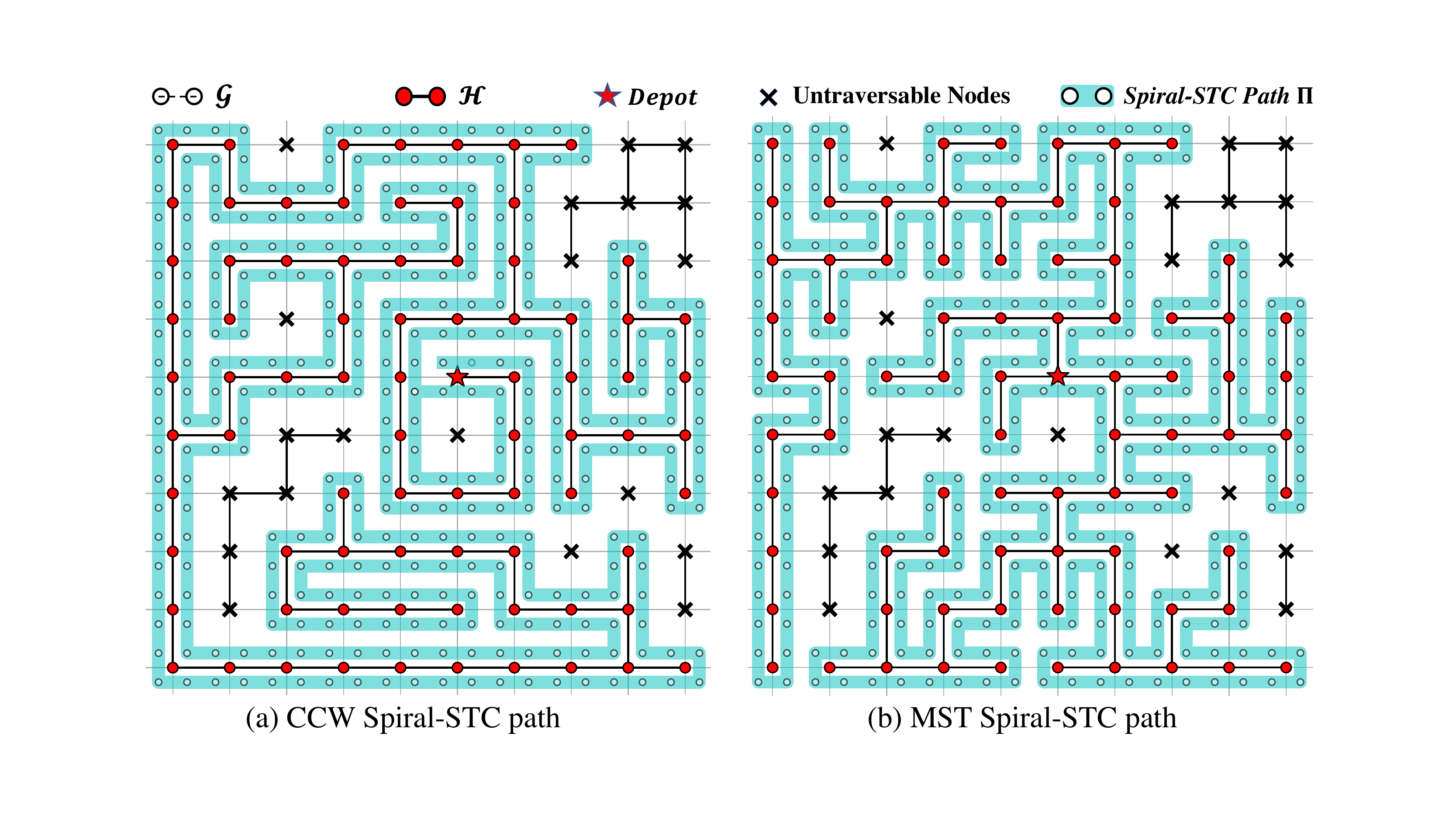}
\caption{Coverage path generated using Spiral-STC in counter-clockwise manner (left) and further improved using MST (right).}
\label{fig:spiral_stc}
\end{figure}

%\begin{algorithm}
%\label{alg:Spiral-STC}
%\SetAlgoLined
%\DontPrintSemicolon
%\SetKwFunction{FMain}{Spiral-STC}
%\SetKwFunction{FTraverse}{Traverse}
%\SetKwProg{Fn}{Function}{:}{}
%\KwIn{$S:$ initial spanning node, $\mathcal{H}:$ spanning graph}
%\KwOut{$\Pi$: Spiral-STC coverage path on $\mathcal{G}$}
%\Fn{\FMain{$S$, $\mathcal{H}$}}{
%    $N_{seen}\leftarrow Set\{S\}$, $E_{seen}\leftarrow Set\{\}$\;
%    $Traverse\,(null,\,S,\,False)$\;
%}
%\Fn{\FTraverse{$p$, $x$, $backtracking$}}{
%    $N_{seen}\leftarrow\{x\}\cup N_{seen}$\;
%    \If{not backtracking}{
%        $E_{seen}\leftarrow \{(p, x)\}\cup E_{seen}$\;
%    }
%    $N_{ccw}\leftarrow$ neighbors of $x$ starting from $p$\;
%    \For{$y: N_{ccw}$}{
%        \If{$y \notin N_{seen}$}{
%            $edge\leftarrow$ spanning-edge ($x$, $y$)\;
%            $motion\leftarrow$move along right-side of $edge$\;
%            $\Pi\leftarrow\Pi\cup\{motion\}$\;
%            $Traverse\,(x,\,y,\,False)$\;
%        }
%    }
%    $N_{bk}\leftarrow$ neighbors of $p$ where $(n, p) \in E_{seen}$\;
%    \For{$z: N_{bk}$}{
%        $motion\leftarrow$move along right-side from $p$ to $z$\;
%        $\Pi\leftarrow\Pi\cup\{motion\}$\;
%        remove $(p, z)$ from $E_{seen}$\;
%        $Traverse\,(p,\,z,\,True)$\;
%    }
%}
%
%\caption{Spiral Spanning Tree Coverage}
%\end{algorithm}
The original Spiral-STC algorithm does not consider the impact of edge weights on the final coverage cost $\mathcal{W}$. Minimal-Spanning-Tree (MST) generated a minimal-weighted coverage path on $\mathcal{H}$ and $\mathcal{G}$ by dropping off the edges with high weights. Fig.~\ref{fig:spiral_stc} gives an example to show coverage planning on a $10\times10$ regular grid graph using standard Spiral-STC and an improvement by MST. Their accumulating costs are $\mathcal{W}=227.02$ and $\mathcal{W}=207.20$, respectively.

\section{Terrain Traversability Map} \label{sec:filtering}
\begin{figure}[!thb]
\centering
\includegraphics[width=0.85\linewidth]{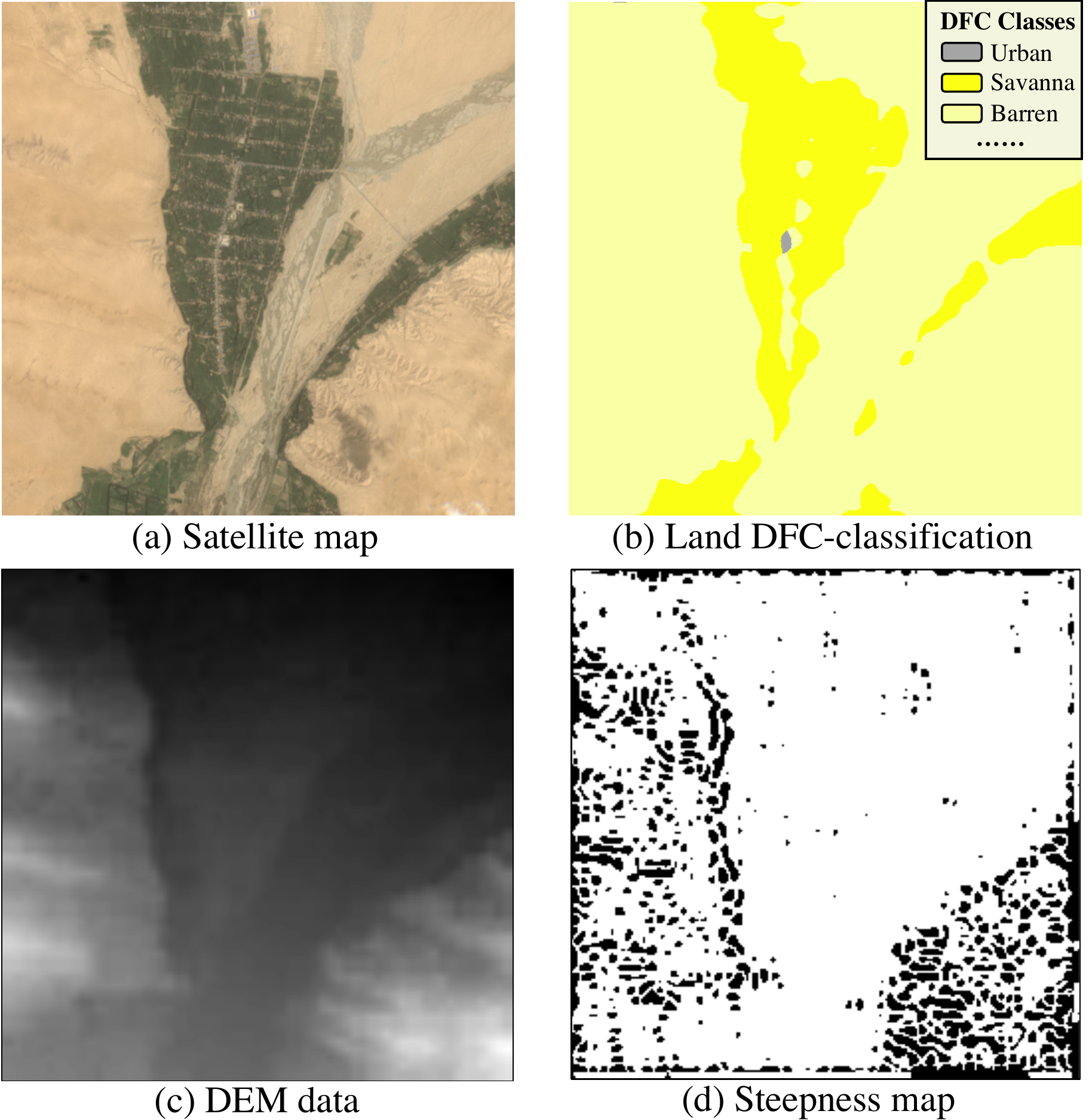}
\caption{Traversability map for a terrain at location ($77.88^\circ$E, $37.35^\circ$N). (a) The original satellite map; (b) Land DFC-classification (only non-plantation regions are displayed); (c) The DEM data representing height information; (d) Steepness traversability map (steepness threshold is set to $25^{\circ}$). Traversable/non-traversable regions resulted from steepness analysis, are highlighted in white and black, respectively.}
\label{fig:terrain_filter}
\end{figure}

A given terrain needs to be processed to obtain traversability map and then generate covering graph, in which the robots can freely move without worrying clear obstacles. Here, we use both digital elevation model (DEM)~\cite{SRTM} and satellite map~\cite{Sentinel} to perform reliable traversability assessment. The DEM is often obtained by the aerial photogrammetric reconstruction. The use of DEM allows us to perform an analysis
of the geometric properties of the given terrain, like terrain slope and height continuity/discontinuity.
Fig.~\ref{fig:terrain_filter} shows an example of DEM data and satellite map.

In our traversability assessment, three main aspects considered herein are
\begin{itemize}
\item Steep regions are unreachable or have the high risk of turnover and sliding for ground mobile robots. Therefore, a slope threshold $25^{\circ}$ is specified based on our preliminary experiments.
\item Isolated regions must be removed from traversability map since they are unreachable from any robot depot.
\item Non-working regions must not be included in traversability map. These regions include forest, shrub-land, marsh, lake, etc.
\end{itemize}

%, results of (b)(d) are further fused into a weighted graph as the area to be covered.

Given a terrain and its initial map (graph), we examine all the edges and remove those edges with the slope greater than a specified threshold (e.g. $25^{\circ}$). The goal is to avoid steep paths as well as obstacles. Then, we remove all unconnected sub-graphs or nodes from the initial graph. As a result, we obtain a traversability map including reachable nodes and their traversal costs. For a given edge $e$, its cost is defined by the normalized slope $\hat{\theta}_e$, and the edge weight contributed by distance $\lVert e \rVert$. That is,
\begin{align}\label{eq:weightcontribution}
    w_e =\alpha\cdot\lVert e \rVert + \beta\cdot \hat{\theta}_e,\;\; e\in\mathcal{G}\; \mathrm{or}\;\mathcal{H},
\end{align}
where $\alpha$ and $\beta$ are the coefficients distinguishing distance contribution and slope contribution. These two parameters can be tuned with the consideration of the significance of distance traversal and the risk of sliding along slopes. Moreover, $\hat{\theta}_e$ is defined as
\begin{align}
\hat{\theta}_e = \frac{\theta_e-\theta_{min}}{\theta_{max}-\theta_{min}},
\end{align}
where $\theta_{max}$ and $\theta_{min}$ is the maximum and minimum slope. Note that $\theta_{max}$ is the slope threshold (e.g. $25^{\circ}$) used in steepness analysis.

%Together with the edge weight described in Section~\ref{sec:problem} and  Fig.~\ref{fig:demo_edges}, the final cost of edge $e$ in the covering graph $\mathcal{G}$ and $\mathcal{H}$ is computed by
%\begin{align}
%    w_e =\alpha\cdot\lVert e \rVert + \beta\cdot \hat{\theta}_e,\;\; e\in\mathcal{G}\; \mathrm{or}\;\mathcal{H},
%\end{align}
%where $\alpha$ and $\beta$ are the coefficients of distance contribution and slope contribution in $w_e$.
%These two parameters can be tuned with the consideration of the significance of distance traversal and the risk of sliding along slopes.

To classify workable and non-workable regions in the satellite map, we use a DNN-based pixel-wise imagery segmentation technique. More specifically, we use the DeepLab Neural Network structure suggested in~\cite{deeplab-origin,deeplab} and apply it to the SEN12M and DFC2020 datasets~\cite{SEN12MS,dfc-2020}.

An example is given in Fig.~\ref{fig:terrain_filter}. The original satellite map is given in Fig.~\ref{fig:terrain_filter}-(a). The traversable regions are shown in Fig.~\ref{fig:terrain_filter}-(b). The DEM data is given Fig.~\ref{fig:terrain_filter}-(c). Its steepness analysis and the results are given in Fig.~\ref{fig:terrain_filter}-(d).

With the consideration of three aspects mentioned above, the two traversability maps are merged into one. Then covering graph $\mathcal{G}$ and spanning graph $\mathcal{H}$ are constructed for these maps.

\section{Our Algorithm} \label{sec:MCPP}

%Here we first introduce our preliminary algorithm namely MSTC${}^\ast$ (MSTC-star) based on MST Spiral-STC, to solve the mCPP problem on weighted graph without material capacity constraint, and then extends the MSTC${}^\ast$ method with limited material capacity constraint.
\subsection{Coverage Path Partition}

\begin{figure}[h]
\centering
\includegraphics[width=\linewidth]{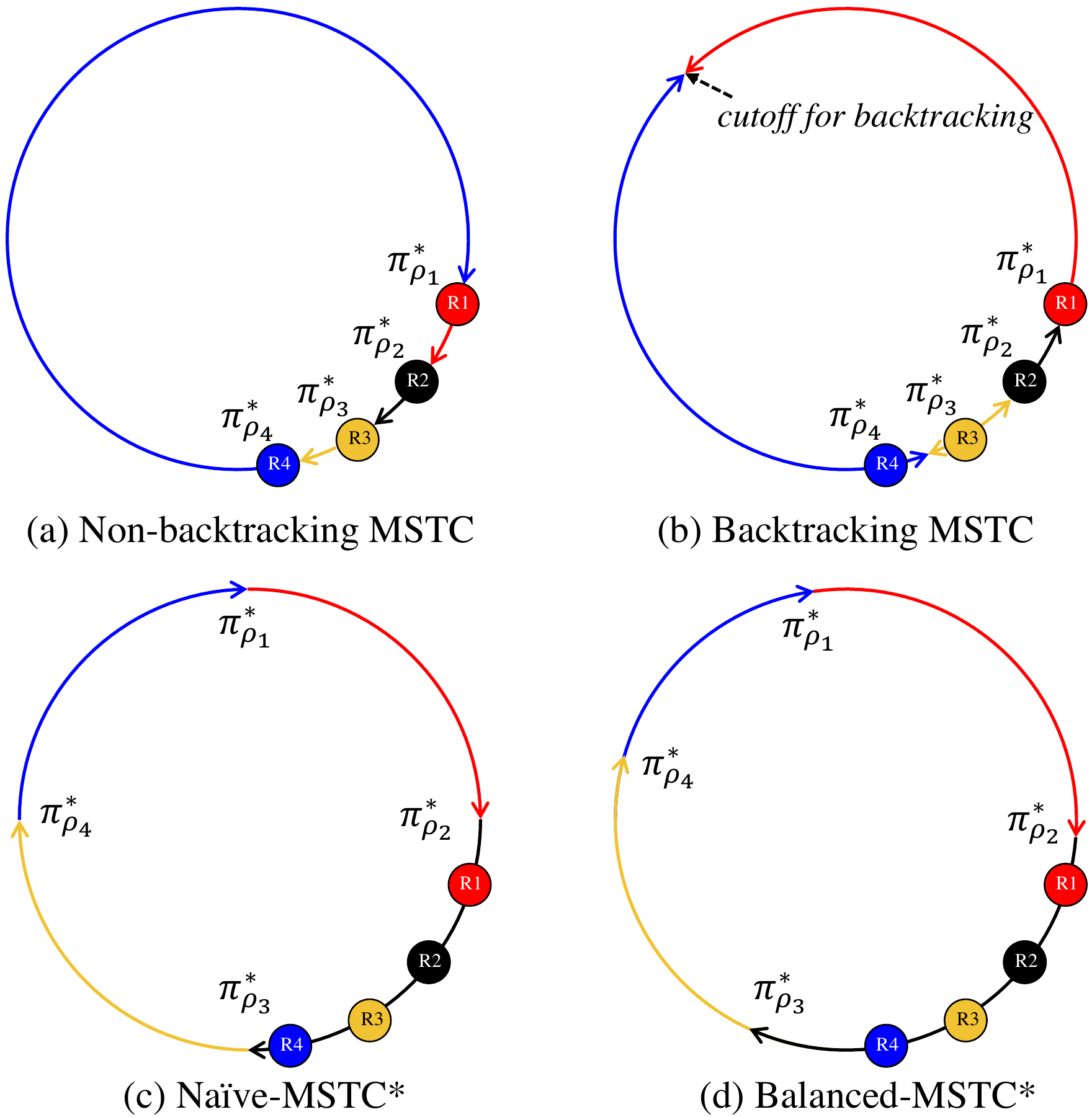}
\caption{Coverage path partition strategies. (a) MSTC; (b) Backtracking MSTC; (c) Na\"ive-MSTC${}^\ast$; (d) Balanced-MSTC${}^\ast$.}
\label{fig:stc_path_partitioning}
\end{figure}

The coverage path planning for multiple robots can be treated as the problem of partitioning a topological loop (see Fig.~\ref{fig:stc_path_partitioning}).
Inspired by MSTC, our algorithm MSTC${}^\ast$ aims at partitioning the entire coverage path $\Pi$ into $k$ partitions and assign each partition to one robot.

\begin{figure*}[hbt!]
\centering
\includegraphics[width=0.85\linewidth]{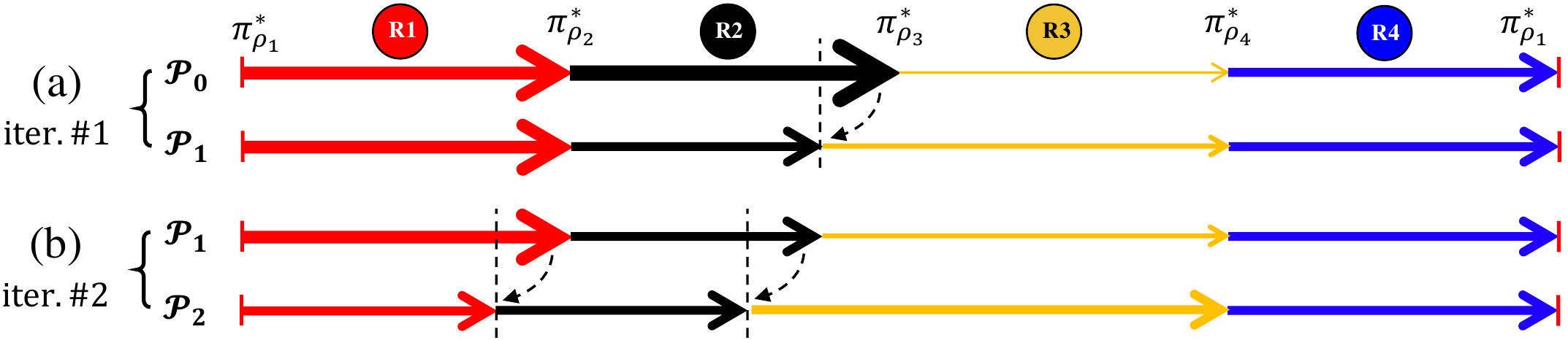}
\caption{Balanced-Cut Partition. (a) Given an initial partitions $\mathcal{P}_0$ generated by Na\"ive-MSTC${}^\ast$, we obtain $\Pi_{min}=\Pi_3$ and $\Pi_{max}=\Pi_2$. (b) Starting with $\mathcal{P}_1$ resulting from the first iteration, we have $\Pi_{min}=\Pi_3$ and $\Pi_{max}=\Pi_1$, and their in-between partition $\Pi_2$.}
\label{fig:demo_find_balanced_cut}
\end{figure*}
$k$ partitions of $\Pi$ can be denoted by $k$ key nodes
\begin{equation}
\mathcal{P}=\{\pi^*_{\key_1}, \pi^*_{\key_2}, \; ... \; ,\pi^*_{\key_k}\}
\end{equation}

Here, we use $\pi^*$ to represent a key node that separates two adjacent partitions. $\key_i$ is the node index from which a partition of coverage path starts. Then a partition is the coverage path $\Pi_i$ consisting of all nodes between two key nodes.
\begin{equation}\label{eqn:coverage_path}
\Pi_i = \{\pi^*_{\key_i}, \pi_{\key_i+1}, \pi_{\key_i+2}, \;...\; , \pi^*_{\key_{i+1}-1}\}
\end{equation}
The main challenge of partitioning $\Pi$ is to choose the key nodes. Non-backtracking MSTC use the depot to be the key nodes~\cite{MSTC}. Each robot moves along its coverage path $\Pi_i$ until it reaches its neighbor robot's depot, as shown in Fig.~\ref{fig:stc_path_partitioning}-(a). This partitioning strategy is inefficient due to the unbalanced workload. Even though backtracking optimization is suggested in~\cite{MSTC}, the resulting coverage path is still uneven. As shown in Fig.~\ref{fig:stc_path_partitioning}-(b), most of coverage paths are executed by robots $\mathcal{R}_1$  and $\mathcal{R}_4$.

\subsection{Our MSTC${}^\ast$ Algorithm}
To  solve  the  problem  of  unbalanced workload exhibiting in the partition strategy of MSTC, we proposed an improved MSTC, namely MSTC${}^\ast$.

\textbf{Na\"ive-MSTC${}^\ast$}:
A straightforward solution is to generate $k$ partitions, yielding the same amount of covering nodes in all the coverage paths $\Pi_i$ (see Fig.~\ref{fig:stc_path_partitioning}-(c)). However, due to the lack of the coverage path costs, the resulting tasks for $k$ robots may be still unbalanced.

\textbf{Balanced-MSTC${}^\ast$}:
Here, we propose a balanced MSTC${}^\ast$ algorithm, aiming at finding a set of well-balanced partitions $\mathcal{P}$ for $\Pi$ with the consideration of coverage path costs. 
Given an initial set of workload partitions, our goal is to generate
balanced workload for all robots by iteratively minimizing
the maximum weights (refer to Eq. (1)).
We elaborate our algorithm as follows.

%on the basis of Naïve-MSTC*.

%, such an even partition can be described as following:
%\begin{equation}
%\mathcal{P}_{Naive-MSTC*}=\{\pi^*_{\rho_i} = \pi_{i\cdot\lceil{\lVert\Pi\rVert/k}\rceil}\,|\,i\in[1, k]\}
%\end{equation}

%It is obvious that Naïve-MSTC* $k$ generates coverage path segments of same number of covering nodes, but maybe each of different coverage cost on $\mathcal{G}$, such a even partition strategy still leaves space to optimize.

%The balanced path partition strategy aims to find a set of well-balanced partition nodes $\mathcal{P}$ on $\Pi$, to further balances the workload of each robot and minimizes the max weights $\mathcal{W}$ on the basis of Naïve-MSTC*.

%As shown in Alg.~\ref{alg:balanced_mstc*_path_partition}, the algorithm starts with the solution of Naïve-MSTC*, i.e. $\mathcal{P}_{Naive-MSTC*}$, and keeps optimizing the partition nodes in a number of iterations.
%For clarity,
%=\{\pi^*_{\key_1}, \pi^*_{\key_2}, \; ... \; ,\pi^*_{\key_k}\}
Given an initial partition $\mathcal{P}_0$ generated using Na\"ive-MSTC${}^\ast$, the coverage path with the maximum weight is denoted by $\Pi_{max}$ and it is partitioned by key node $\pi^*_{\rho_{max}}$. Analogically, the coverage path with the minimum weight is denoted by $\Pi_{min}$ and it is partitioned by key node $\pi^*_{\rho_{min}}$.
An example is shown in Fig.~\ref{fig:demo_find_balanced_cut}. For simplicity, we unfold a loop (i.e., the loop in Fig.~\ref{fig:stc_path_partitioning}) into line segments. An initial partition is $\mathcal{P}_0 = \{\pi^{\ast}_{\key_1}, \pi^{\ast}_{\key_2}, \pi^{\ast}_{\key_3}, \pi^{\ast}_{\key_4}\}$ for four robots $\mathcal{R}_1$, $\mathcal{R}_2$, $\mathcal{R}_3$ and $\mathcal{R}_4$. Even though the nodes are evenly distributed, their weights are unbalanced.
Here, a higher weight corresponds to a thicker line segment. As shown in Fig.~\ref{fig:demo_find_balanced_cut}-(a), the coverage path of $\mathcal{R}_2$ has the maximum weight. That is, $\Pi_{max}=\Pi_2$. The coverage path of $\mathcal{R}_3$ has the minimum weight. Therefore, $\Pi_{min}=\Pi_3$. If $\Pi_{max}$ and $\Pi_{min}$ are adjacent, we iteratively remove the nodes at their boundary from $\Pi_{max}$ and append them into $\Pi_{min}$ until they are balanced. However, if $\Pi_{max}$ and $\Pi_{min}$ are not adjacent.  We gradually shift the nodes from $\Pi_{max}$ to $\Pi_{min}$ through in-between partitions. An example is given in  Fig.~\ref{fig:demo_find_balanced_cut}-(b), in which $\Pi_{max}=\Pi_1$ and $\Pi_{min}=\Pi_3$. In Fig.~\ref{fig:demo_find_balanced_cut}-(b), the in-between partition is $\Pi_{2}$. In other words, we remove the rightmost nodes from $\Pi_1$ and append them to the in-between partition $\Pi_{2}$. Meanwhile, we remove the rightmost nodes from $\Pi_2$ and append them to the partition $\Pi_{3}$. During the shifting, the workload of the in-between partitions remain unchanged.

To determine the potential shift partition nodes, a straightforward strategy is to linearly search $\Pi_{min}$ and $\Pi_{max}$. However, linear search can be very computationally expensive. Therefore, we propose the strategy of \emph{balanced cut} to accelerate the search. This strategy relies on binary search by iteratively updating search bound. In fact, finding the best partitions is a NP-hard problem. However, our strategy is a greedy algorithm to gradually approximate the optimal partitions. 
Our \emph{balanced cut} algorithm is summarized as the pseudo-code in Algorithm~\ref{alg:balanced_cut}.
The Balanced-MSTC${}^\ast$ algorithm is summarized as the pseudo-code in Algorithm~\ref{alg:balanced_mstc*}.

\begin{algorithm}
\label{alg:balanced_cut}
\SetAlgoLined
\DontPrintSemicolon
\caption{Balanced-Cut ($\mathcal{P}$, $\Pi_{min}$, $\Pi_{max}$)}
% \SetKwFunction{FFind}{Balanced-Cut}
% \SetKwProg{Fn}{Function}{:}{}
\KwIn{A set of partitions $\mathcal{P}$, and $\Pi_{min}$ and $\Pi_{max}$}
\KwOut{A set of balanced partitions}
% \Fn{\FFind{$\mathcal{P}$,  $\Pi_{min}$, $\Pi_{max}$}}{
% $\pi^*_{\rho_{min}}\leftarrow$ correspondent partition node of $\Pi_{min}$\;
% $\pi^*_{\rho_{max}}\leftarrow$ correspondent partition node of $\Pi_{max}$\;
$\mathrm{left}\leftarrow 0$\;
$\mathrm{right}\leftarrow\lVert\Pi_{min}\rVert + \lVert\Pi_{max}\rVert$\;
$m=\lfloor\frac{\mathrm{left}+\mathrm{right}}{2}\rfloor$\;
\While{$\mathrm{left} < \mathrm{right}$}
{
    $s \leftarrow \left(\lfloor\frac{\mathrm{left}+\mathrm{right}}{2}\rfloor - m\right)$\;
    $m \leftarrow \lfloor\frac{\mathrm{left}+\mathrm{right}}{2}\rfloor$\;
    \For{$\Pi_i: \Pi_{min}, \Pi_{{min}+1}, , ..., \Pi_{max}$}{
        shift partition node $\pi^{\ast}_{\rho_i}$ of $\Pi_i$ by $s$ nodes\;
        update the coverage cost $\mathcal{W}_{\Pi_i}$ for $\Pi_i$\;
    }
    update $\mathcal{P}$\; %if $\max_{1\leq i\leq k}\nolimits \left(\mathcal{W}_{\Pi_i}\right)$
    %is smaller\;
    \eIf{$\mathcal{W}_{\Pi_{min}} < \mathcal{W}_{\Pi_{max}}$}{
        $\mathrm{left} \leftarrow (m + 1)$
    }
    {
        $\mathrm{right} \leftarrow  (m - 1)$
    }
}
    \Return{$\mathcal{P}$}\;
% }
\end{algorithm}

\begin{algorithm}
\label{alg:balanced_mstc*}
\SetAlgoLined
\DontPrintSemicolon
\caption{Balanced-MSTC${}^\ast$($\mathcal{G}$, $\mathcal{H}$, $k$)}
\SetKwFunction{FMain}{Balanced-MSTC${}^\ast$}
% \SetKwFunction{FFind}{Balanced-Cut}
\SetKwProg{Fn}{Function}{:}{}
\KwIn{Covering graph $\mathcal{G}$,  spanning graph $\mathcal{H}$ and $k$ robots}
\KwOut{$k$ coverage paths $\{\Pi_1, \Pi_2, ..., \Pi_k\}$}
% \Fn{\FMain{$\mathcal{G}$, $\mathcal{H}$, $k$}}{
$\Pi \leftarrow$ Spiral-STC path for a single robot using $\mathcal{G}$, $\mathcal{H}$\;
$\mathcal{P}\leftarrow
    \{\pi^{\ast}_{\rho_i}\}$\;
determine $\Pi_{max}$ and $\Pi_{min}$\;
compute their weights $\mathcal{W}_{\Pi_{max}}$ and $\mathcal{W}_{\Pi_{min}}$\;
    %= \pi_{i\cdot\lceil{\lVert\Pi\rVert/k}\rceil}\,|\,i\in[1, k]\}$\;
    \While{$\mathcal{W}_{\Pi_{max}} > \mathcal{W}_{\Pi_{min}}$}{
        \For{$\Pi_i: \Pi_1, \Pi_2, , ..., \Pi_k$}{
        compute weights $\mathcal{W}_{\Pi_i}$\;
        $\mathcal{W}_{\Pi_{max}} = \max(\mathcal{W}_{\Pi_{max}}, \mathcal{W}_{\Pi_i})$\;
        $\mathcal{W}_{\Pi_{min}} = \min(\mathcal{W}_{\Pi_{min}}, \mathcal{W}_{\Pi_i})$\;
    }
        %compute weights $\mathcal{W}_{i}$\;
        %find $\mathcal{W}_{\max}$ and $\mathcal{W}_{\min}$\;

        $\Pi_{max}\leftarrow$ coverage path associating with $\mathcal{W}_{\Pi_{max}}$\;
        $\Pi_{min}\leftarrow$ coverage path associating with $\mathcal{W}_{\Pi_{min}}$\;
        $\mathcal{P} \leftarrow$ Balanced-Cut$\,(\mathcal{P},\,\Pi_{min},\,\Pi_{max})$\;
}
% }
\Return $\Pi$= $\{\Pi_1, \Pi_2, ..., \Pi_k\}$ partitioned by $\mathcal{P}$ (Eq.~\ref{eqn:coverage_path})\;
\end{algorithm}
%However, linear searching on $\mathcal{S}$ yields to enormous calculation of shortest path and the correspondent cost, as described above.
%In order to speedup the search, our implementation of \textit{Balanced-Cut} in Alg.~\ref{alg:balanced_cut} exploits a binary search to reduce the time complexity, line 13 and 15 shows the update conditions of left and right bound of binary search.
%While such a binary search scheme does not guarantee an optimal solution which can be found by linear search for the completeness is not guaranteed, it can still find good solution of balanced path partition node sets for further multi-robot planning within acceptable computation time.

\subsection{MSTC${}^\ast$ under Limited Workload Capacity}
In this section, we explain the process of extending Balanced-MSTC${}^\ast$ algorithm to handle limited workload capacity. Assume that robots have equal workload capacity $c$. Due to limited workload capacity, a robot needs to return to its depot for refilling. Let the traversal time spending on going back to depots and refilling be  $r_i=\lceil\frac{\lVert\Pi_i\rVert}{c}\rceil$ for robot $\mathcal{R}_i$. Therefore, the total time spending on returning and refilling is $\sum_{i=1}^{k}{r_i}$.

If the capacity $c\ge \frac{\lVert\Pi\rVert}{k}$, $\forall i\in[1, k]$, we have $r_i=1$. We can obtain $k$ partitions without refilling. %using  Alg.~\ref{alg:balanced_mstc*} as discussed above in Sec.~\ref{sec:MCPP}.
If the capacity $c<\frac{\lVert\Pi\rVert}{k}$, every robot needs to refill in order to accomplish the tasks. This will have additional $n=\sum_{i=1}^{k}{r_i}-k$ partition nodes in $\Pi$, which requires to be evenly distributed to $k$ robots. More specifically, we can perform Balanced-MSTC${}^\ast$ by assuming $n$ robots.  After obtaining $n$ partitions, we merge some adjacent partitions to generate $k$ new partitions for $k$ robots
\begin{equation}
\Pi_i=\bigcup_{j=i\cdot\frac{n}{k}}^{(i+1)\cdot\frac{n}{k}-1}\nolimits \Pi'_j.%,\;\;i\in[1, k]
\end{equation}

\section{Results \& Analysis} \label{sec:exp}
We will now explain some implementation details of our algorithms and show our experimental results. Moreover, we compare our algorithms (Na\"ive-MSTC${}^\ast$ and Balanced-MSTC${}^\ast$) against existing spiral-STC based methods: classic MSTC (MSTC-NB)~\cite{MSTC}, MSTC with backtracking (MSTC-BO)~\cite{MSTC} and Multi-robot Forest Coverage (MFC)~\cite{MFC}.

\subsection{Implementation}
We used DEM data from SRTM Digital Elevation Database of CGIAR\cite{SRTM} for steepness anaylysis and satellite map data from Sentinel-2 imagery of ESA\cite{Sentinel} for DFC-classification. Then we incorporate them into the original terrain map to generate two traversability maps. In Eq.~\ref{eq:weightcontribution}, we set $\alpha=\frac{1}{3}$ and $\beta=\frac{2}{3}$.

\subsection{Regular Grid Map}
\begin{figure}[!htb]
\centering
\includegraphics[width=\linewidth]{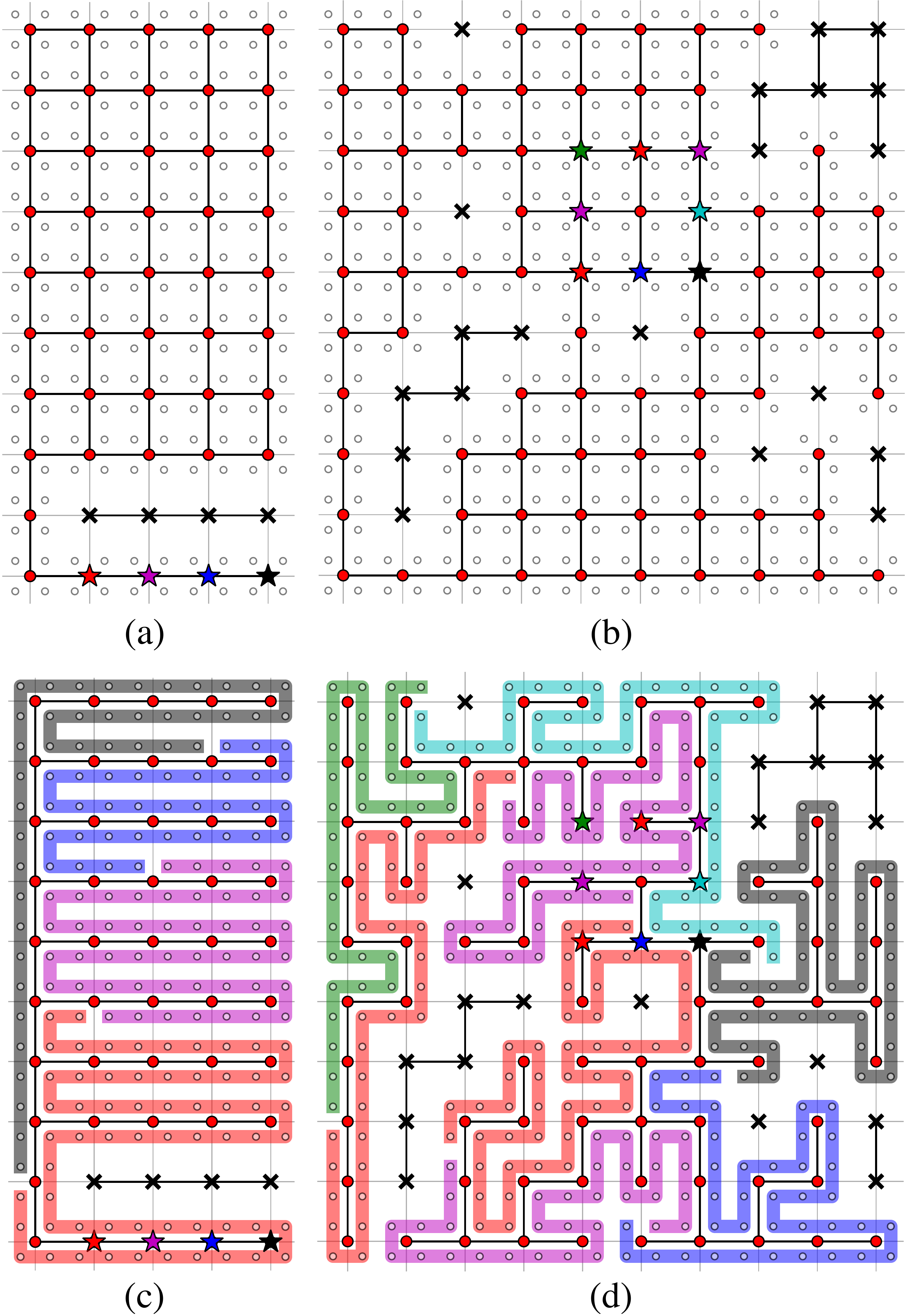}
\caption{Regular grid terrains. Top: terrains to be covered; Bottom: mCPP using our Balanced-MSTC${}^\ast$. (a)(c) k=4 and c=$\infty$. (b)(d) k=8 and c=$\infty$.}
\label{fig:artificial_scenes}
\end{figure}

We first tested our algorithm on small regular grid maps (see Figs.~\ref{fig:artificial_scenes}-(a) and (b)).
The first regular grid map, shown in Fig.~\ref{fig:artificial_scenes}(a), is a unweighted blocked terrain used in~\cite{MFC, MFC-weighted, MFC-2010}.
Four robot depots are represented by stars and located at the lower-left cells.
All four robots start from their individual depot. The second scenario, shown in Fig.~\ref{fig:artificial_scenes}-(b), is a randomly generated $10\times10$ weighted terrain, with 8 robot depots and some random blocked cells.
The blocked cells are indicated by $\mathbf{\times}$ in these figures. The results of coverage paths are given in Figs.~\ref{fig:artificial_scenes}-(c) and (d).

%We first use two smaller artificial terrains as test cases to compare these methods. The first one refers to a unweighted blocked terrain used in~\cite{MFC, MFC-weighted, MFC-2010}.
%Fig.~\ref{fig:artificial_scenes}(a) depicts the blocked terrain with 4 depots, where robots can only reach the most parts of the terrain via a narrow passageway, more detailed discussions of this scenario can be found in~\cite{MFC}.
%The second test case is a $10\times10$ random generated weighted spanning graph with 8 depots, as depicted in Fig.~\ref{fig:artificial_scenes}(b).

\begin{figure}[!htb]
\centering
\includegraphics[width=\linewidth]{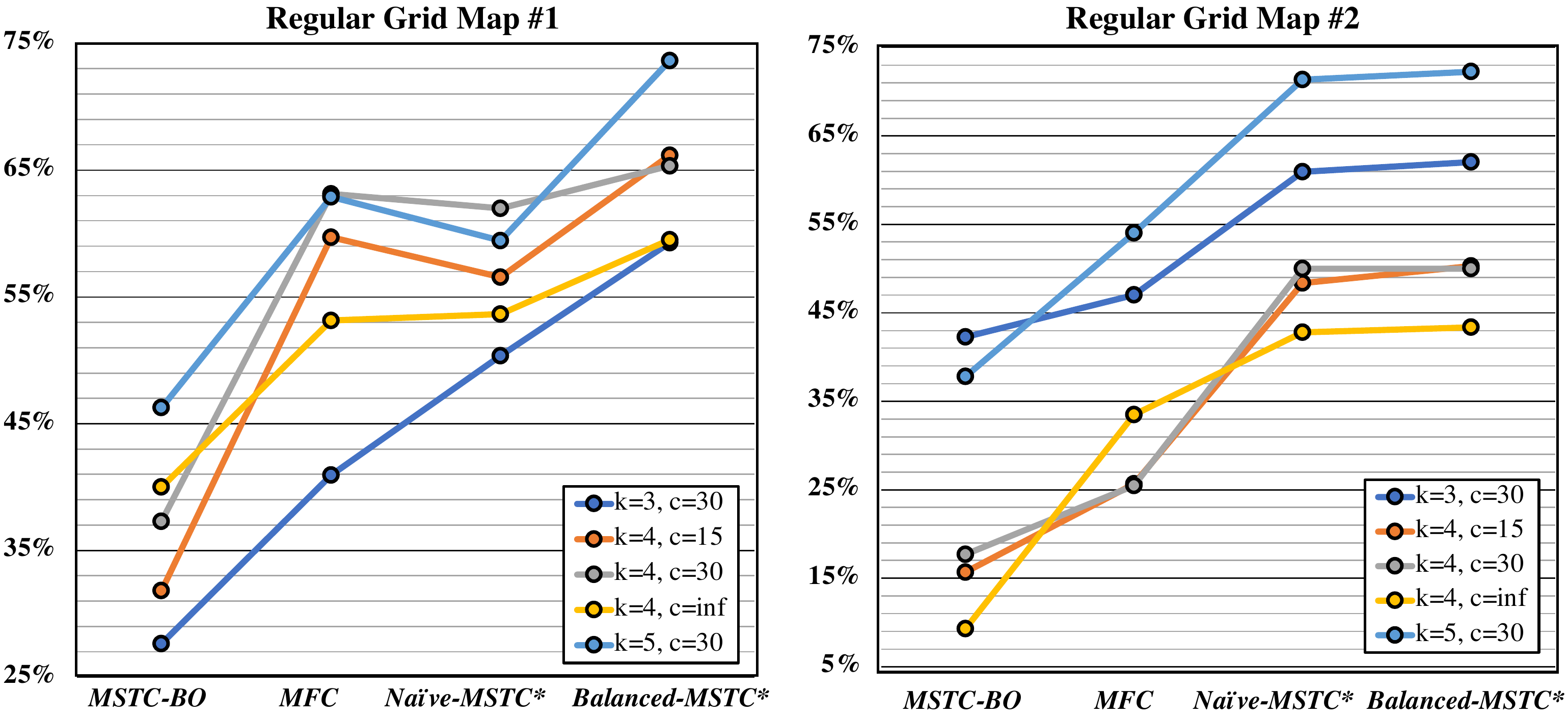}
\caption{Performance and scalability comparison for regular grid terrains in terms of the reduction ratio of the maximum weights.}
\label{fig:artificial_scene_result}
\end{figure}

\textbf{Comparison}: Fig.~\ref{fig:artificial_scene_result}
shows the performance scalability of a few spiral-STC based methods
with respect to the number of robots $k$ and workload capacity $c$. We evaluated the performance improvement using the reduction ratio of the maximum weights (refer to Eq.~(\ref{eqn:mcpp_task})). We also compared our MSTC${}^\ast$ against MSTC-BO and MFC. our Balanced-MSTC${}^\ast$ algorithm outperforms others. Especially, for the first terrain, Balanced-MSTC${}^\ast$ has significant improvement against Na\"ive-MSTC${}^\ast$.

%shows the comparisons of max weights between 5 methods in the two artificial terrain test cases motioned above.
%Different number of robots (denoted as $k$) and capacities (denoted as $c$) are combined, to compare the efficiency of each methods.
%As we can see, the Balanced-MSTC* method generates better coverage plans with smallest max weights in both test cases.
%While it improves a lot from Naïve-MSTC* on the first blocked terrain test case, it improves few from Naïve-MSTC* on the second random terrain test case.
%The reason of different degrees of improvement between Naïve-MSTC* and Balanced-MSTC* in this case lies on that the more the covering nodes are evenly distributed, the less space of optimizing the balance of workload by performing balanced path partition.

\subsection{Field Terrain}
\begin{figure}[!htb]
\centering
\includegraphics[width=0.9\linewidth]{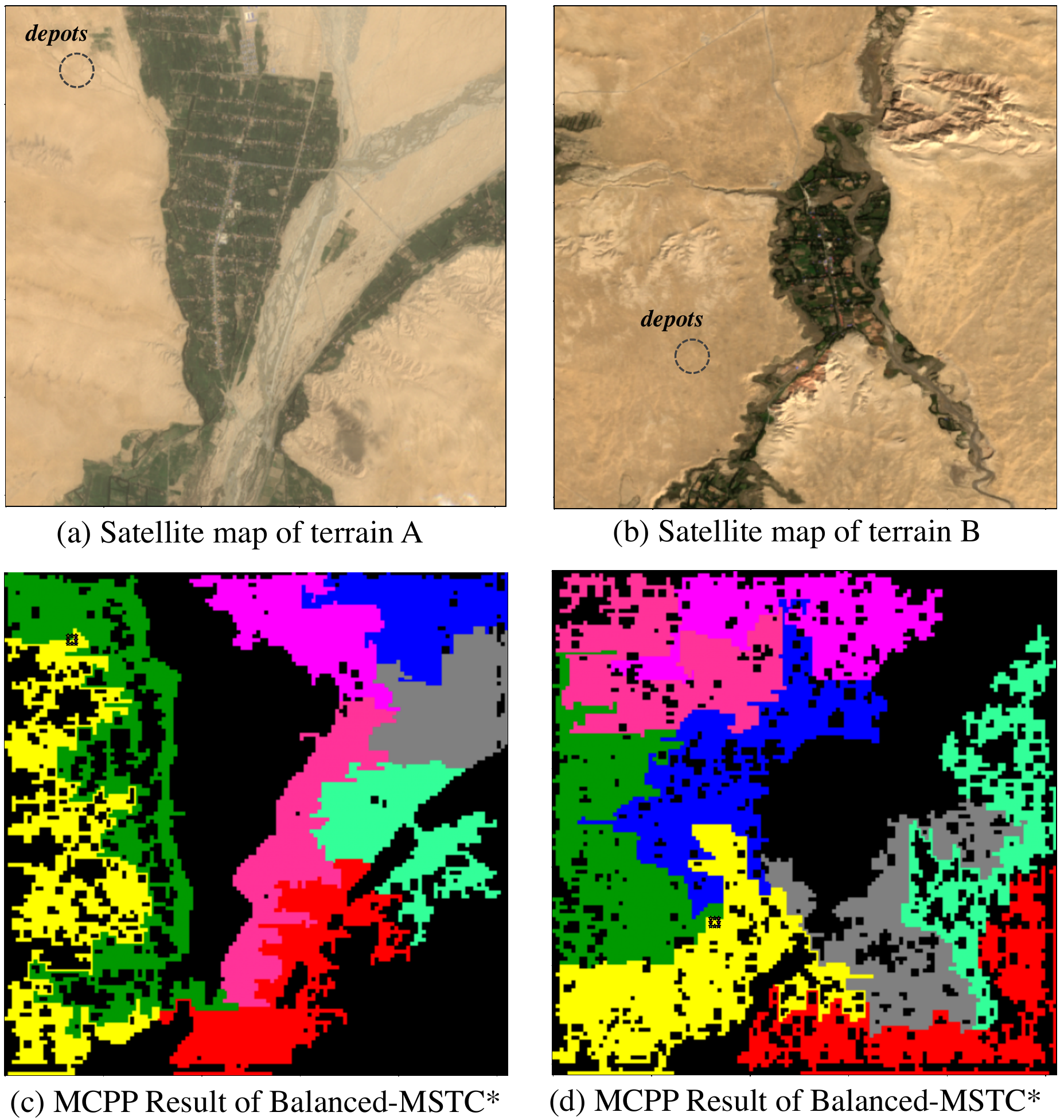}
\caption{Real Terrains. The robot depots are circled in the given map. (a) Satellite map at ($77.88^\circ$E, $37.35^\circ$N); (b) Satellite map at ($78.06^\circ$E, $37.26^\circ$N); (c)(d) The mCPP results using our Balanced-MSTC${}^\ast$ algorithm ($k=8$, $c=400$).}
\label{fig:real_terrain}
\end{figure}

We also apply our algorithm to field terrains in the real-world applications, as shown in Figs.~\ref{fig:real_terrain}-(a) and (b).
The mCPP results using our Balanced-MSTC${}^\ast$ algorithm are given in Figs.~\ref{fig:real_terrain}-(c) and (d), respectively.
The given satellite maps in Figs.~\ref{fig:real_terrain}-(a) and (b) are divided into $256\times256$ cells.
After traversability processing (i.e., DFC-classification and steepness analysis), their spanning graphs include 10238 nodes and 17931 edges in Fig.~\ref{fig:real_terrain}-(a), and 12476 nodes and 19235 edges in Fig.~\ref{fig:real_terrain}-(b).

%use two real terrain test cases to compare the performance of each methods under real scenarios.
%Fig.~\ref{fig:real_terrain} shows the original satellite map ($256\times256$) of terrains, as well as the final weighted graph (each has 10238 nodes, 17931 edges and 10238 nodes, 17931 edges) by fusing the results of land DFC-classification and steepness filtering.

\begin{figure}[!ht]
\centering
\includegraphics[width=\linewidth]{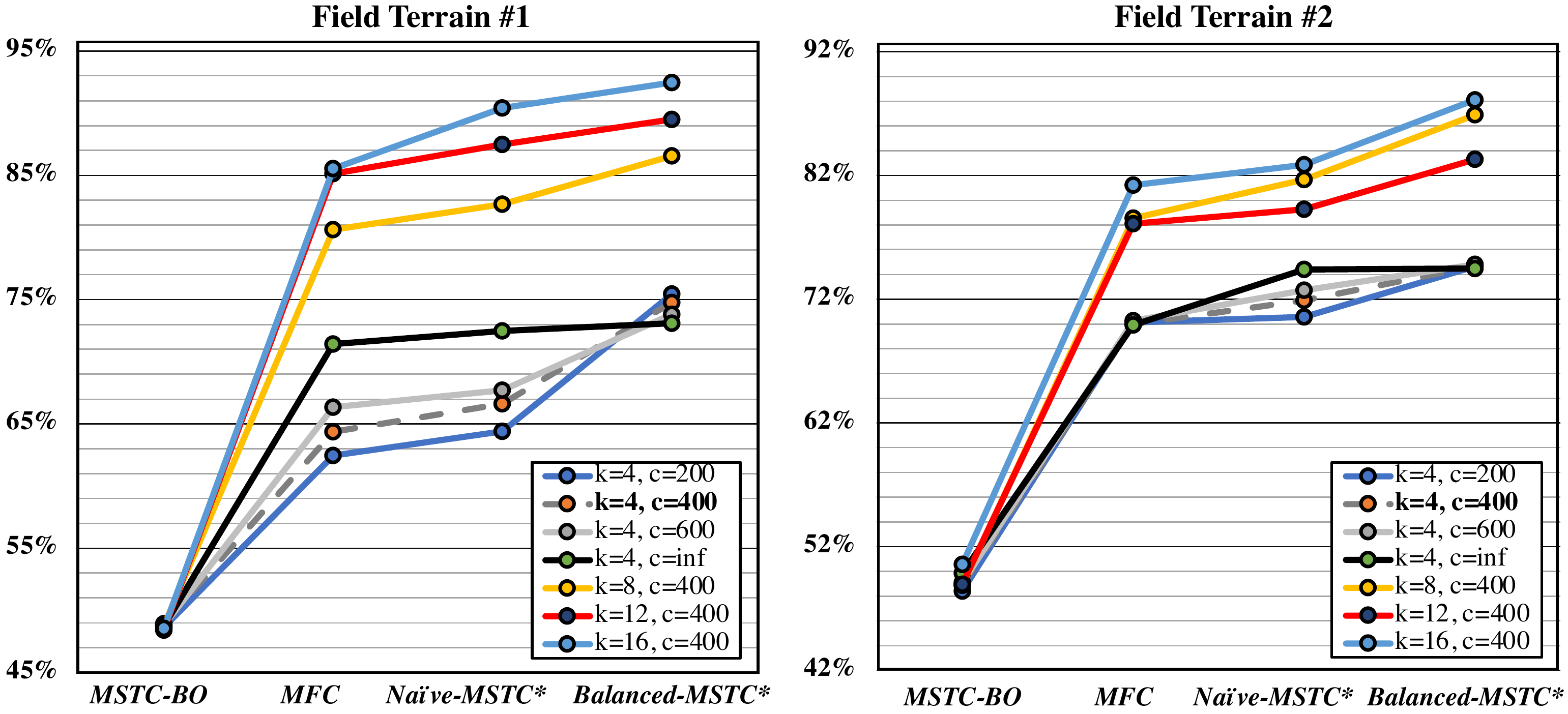}
\caption{Performance and scalability comparison for field terrains in terms of the reduction ratio of the maximum weights.}
\label{fig:real_scene_result}
\end{figure}

\textbf{Comparison}: We compare our algorithms against the spiral-STC based methods in terms of the reduction ratio of the maximum weights. As shown in Fig.~\ref{fig:real_scene_result}, our Na\"ive-MSTC${}^\ast$ and Balanced-MSTC${}^\ast$ outperform MSTC-BO and MFC.

%the max weights reduction ratio compared with MSTC-NB in Fig.~\ref{fig:real_scene_result}.
%Generally speaking, the result shows that Naïve-MSTC* and Balanced-MSTC* outperform other Spiral-STC based methods. Depending on the space to optimize the balance of workload, Balanced-MSTC* improves efficiency less or more compared with Naïve-MSTC*.

In addition, we further compare the scalability
with respect to the number of robots $k$ and workload capacity  $c$. 
As shown in Fig.~\ref{fig:real_scene_result}, MSTC-BO shows the worst scalability. In general, our Na\"ive-MSTC${}^\ast$ and Balanced-MSTC${}^\ast$ exhibit a higher scalability as the number of robots and the workload capacity increase. 
For example, the reduction ratio increases as the number of robots increases ($k=4, 8, 12, 16$ for $c=400$). 
However, we observed that the scalability
of our algorithm is output sensitive. 
The performance improvement becomes less significant as the workload capacity continuously increases. 
For example, the performance of Balanced-MSTC${}^\ast$ becomes similar to Na\"ive-MSTC${}^\ast$ for $k=4, c=\infty$ (see Fig.~\ref{fig:real_scene_result}).
Intuitively, our Balanced-MSTC${}^\ast$ greatly benefits from the shortest path traversal during workload refilling. If refilling is unnecessary (c=$\infty$), the performance gain can be neglected. 

%For example, the performance of Balanced-MSTC${}^\ast$ is similar to Na\"ive-MSTC${}^\ast$ for $k=4, c=\infty$.

%impacts of $k$ and $c$ on the max weights reduction ratio, two control experiments with $k$ and $c$ each setting as the control variable are designed to compare the impact of the other variable.
%As we can see from Fig.~\ref{fig:real_scene_result}(a)(b), for control experiment with $c=400$ on both real terrain test cases, the larger the $k$ value is, the more max weights reduction of MFC, Naïve-MSTC* and Balanced-MSTC*, while the reduction of MSTC-BO keeps around a steady level.
%What's more, \textit{the law of diminishing marginal utility} appears with $k$ value increasing, that is to say, the efficiency improves less with more robots (the reduction ratio increment keeps decreasing among $k=4, 8, 12, 16$ when $c=400$).
%For control experiment with $k=4$ on both terrain test cases as shown in both Fig.~\ref{fig:real_scene_result}(a)(b), we can draw the conclusion that the larger the $c$ value is, the less space of Balanced-MSTC* to optimize the balance of workload (the performance of Naïve-MSTC* is very close to Balanced-MSTC* for $k=4, c=inf$). 
\section{Conclusions \& Future Work}
We have presented an efficient algorithm MSTC${}^\ast$ for multi-robot coverage path planning. Our algorithm improved spiral spanning tree coverage method by incorporating strict physical constraints like terrain traversability and material load capacity. We have performed extensive comparison with other mCPP methods both in regular grid maps and real-world terrains. Our method showed significant performance improvement against existing spiral-STC mCPP methods.

There are a few limitations in our algorithm. 
Our algorithm to find the partitions is greedy such that there is no guarantee to find the best. Our algorithm requires a few problem-dependent parameters such as $\alpha$ and $\beta$ to compute the edge costs in covering graph $\mathcal{G}$ and spanning graph $\mathcal{H}$.

For future work, we would like to apply our techniques to real robots. Since the runtime communication and synchronization can cause overhead, we would like to  investigate the extensions of our algorithm to de-centered environments.

\section*{Acknowledgments}
This work was supported by the NSFC-RGC Joint Program (NSFC61631166002, HKU103/16) and partially supported by Huawei Ascend AI Computing Platform and CANN (Compute Architecture for Neural Networks).

\addtolength{\textheight}{0cm}   % This command serves to balance

% bib %%%%%%%%%%%%%%%%%%%%%%%%%%%%%%%%%%%%%%%%%%%%%%%%%%%%%%%%%%%%%%%%%%%%%%%%%%
% \vfill\pagebreak
\bibliographystyle{IEEEtran}
\bibliography{CPP}

\end{document}